\documentclass[lettersize,journal]{IEEEtran}
\usepackage{amsmath,amsfonts}
\usepackage{algorithm}
\usepackage{array}
\usepackage[caption=false,font=normalsize,labelfont=sf,textfont=sf]{subfig}
\usepackage{textcomp}
\usepackage{stfloats}
\usepackage{url}
\usepackage{verbatim}
\usepackage{graphicx}
\usepackage{cite}
\newcommand{\eg}{\textit{e}.\textit{g}.}
\newcommand{\ie}{\textit{i}.\textit{e}.}
\newcommand{\etal}{\textit{et~al.}}

\usepackage{algpseudocode}
\usepackage{amssymb}
\usepackage{float}
\usepackage{booktabs}
\usepackage{wrapfig}
\usepackage{microtype}
\usepackage{hyperref}

\usepackage[hypcap=false]{caption}

\usepackage[percent]{overpic}
\usepackage{xcolor}
\usepackage{tikz}

\begin{document}

\title{Look, Compare and Draw:  Differential Query Transformer for Automatic Oil Painting}

\author{~Lingyu~Liu,~Yaxiong Wang$^\dag$,~Li Zhu,~Lizi Liao,~Zhedong Zheng$^\dag$
\thanks{Lingyu Liu and Li Zhu are with the School of Software, Xi'an Jiaotong University, Xi'an, 710049, China. (e-mail: liulingyu@stu.xjtu.edu.cn; zhuli@mail.xjtu.edu.cn).}
\thanks{Yaxiong Wang is with the School of Computer and Information Science, Hefei University of Technology, Jianghuai Advance Technology Center, Anhui Provincial Key Laboratory of Humanoid Robots. Anhui Provincial Industry Innovation Center of Humanoid Robots, Hefei, 230000, China. (e-mail: wangyx15@stu.xjtu.edu.cn).}
\thanks{Lizi Liao is with the Singapore Management University, 188065, Singapore (e-mail: lzliao@smu.edu.sg).}
\thanks{Zhedong Zheng is with Faculty of Science and Technology, and Institute of Collaborative Innovation, University of Macau, Macau, 999078, China.   (e-mail: zhedongzheng@um.edu.mo).}
\thanks{$^\dag$ Corresponding author.}	}

\markboth{Journal of \LaTeX\ Class Files,~Vol.~14, No.~8, August~2021}%
{Shell \MakeLowercase{\textit{et al.}}: A Sample Article Using IEEEtran.cls for IEEE Journals}

\maketitle

\begin{abstract}
This work introduces a new approach to automatic oil painting that emphasizes the creation of dynamic and expressive brushstrokes. A pivotal challenge lies in mitigating the duplicate and common-place strokes, which often lead to less aesthetic outcomes.
Inspired by the human painting process, \ie, observing, comparing, and drawing, we incorporate differential image analysis into a neural oil painting model, allowing the model to effectively concentrate on the incremental impact of successive brushstrokes. To operationalize this concept, we propose the Differential Query Transformer (DQ-Transformer), a new architecture that leverages differentially derived image representations enriched with positional encoding to guide the stroke prediction process. 
This integration enables the model to maintain heightened sensitivity to local details, resulting in more refined and nuanced stroke generation. Furthermore, we incorporate adversarial training into our framework, enhancing the accuracy of stroke prediction and thereby improving the overall realism and fidelity of the synthesized paintings. 
Extensive qualitative evaluations, complemented by a controlled user study, validate that our DQ-Transformer surpasses existing methods in both visual realism and artistic authenticity, typically achieving these results with fewer strokes.
The stroke-by-stroke painting animations are available on our project website \footnote{\url{https://differential-query-painter.github.io/DQ-painter/}}.
\end{abstract}


\section{Introduction}
\label{sec:intro}
\IEEEPARstart{P}{ainting} is a common form of human artistic expression, but it requires a certain level of technical skill. Computer-aided art~\cite{hertzmann2025generative,hertzmann2024toward,Hertzmann2024art,Hertzmann2024segmentation,isenberg2006non,kyprianidis2012state,rosin2012image,rosin2022nprportrait} enables people without professional drawing skills to create their own artistic works.
Neural oil painting~\cite{Intelli-Paint,liang2022drawing, wang2023sketchknitter, Intelligent-paint, clipdraw} has emerged as a promising paradigm for artistic image transformation by simulating the brushstrokes of oil paintings through hierarchical stroke rendering. It aims to guide machines in progressively generating images by emulating authentic oil painting brushstrokes, from coarse to fine, on a digital canvas, thereby imparting to the images the characteristic texture of oil paintings. 

\begin{figure*}
\centering
    \includegraphics[width=0.97\textwidth]{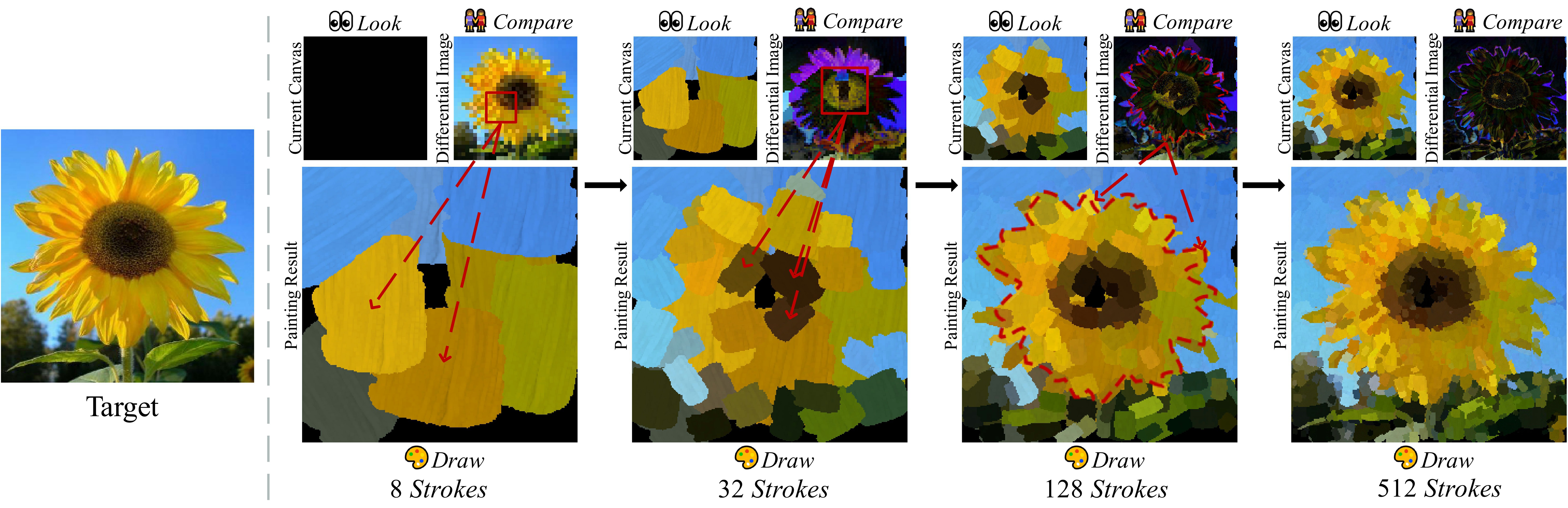}
    \caption{Differential image-guided inference process. We present four intermediate stages of oil painting according to a real target image (left). Each stage is illustrated with a diagram, where the top-left corner shows the current canvas, the top-right corner displays the corresponding differential image for that stage, and the bottom part presents the painting result inferred by our model. We observe that since we explicitly compare the content in the differential images during training, our model tends to add strokes in areas where discrepancies are more pronounced, thereby progressively reducing the discrepancy content within the differential images.}
    \label{fig:di}
\end{figure*}

Traditional stroke-based rendering methods typically rely on step-wise greedy search and heuristic optimization, which often lead to low efficiency~\cite{haeberli1990paint,Litwinowicz_1997,survey_2003,Im2Oil}. As noted by Hu~\etal~\cite{Compositional_Neural_Painter}, deep learning-based methods have gained traction, employing a variety of strategies such as reinforcement learning~\cite{LearningtoPaint,Intelligent-paint,Compositional_Neural_Painter}, neural networks~\cite{PaintTransformer}, and optimization-based approaches~\cite{StylizedNeuralPainting,Parameterized}. 
While these methods have validated promising painting results, challenges in achieving higher efficiency and effectiveness in practical applications persist. For example, Hu~\etal~\cite{Compositional_Neural_Painter} develop a reinforcement learning-based agent trained on real images to dynamically determine the painting sequence, but it struggles with generalization, and becomes unstable when faced with unseen images. 
Similarly, Zou \etal~\cite{StylizedNeuralPainting} introduce a stroke optimization method that achieves high-quality results but requires extremely long inference times. On the other hand,  Liu \etal~\cite{PaintTransformer} directly construct a neural network to efficiently predict a set of strokes. However, this method often produces coarse strokes and particularly fails to capture fine details at the canvas boundaries.

Despite varying learning strategies within specific models, the prevailing works all adhere to the iterative learning paradigm, that is, predicting the subsequent brushstroke based on the current one. In line with this paradigm, existing methodologies employ a rather direct approach by generating the forthcoming brushstroke directly using the existing stroke as input. 
We contend that this predictive approach suffers from the absence of an intermediate guidance from the current stroke to the next, which becomes particularly challenging when there is a significant divergence between the paintings in the early steps of prediction.
Conversely, in the human painting process, artists frequently observe and compare differences between their current work and the target painting before deciding on the subsequent brushwork. Motivated by this procedure, we propose the incorporation of image discrepancy as a form of intermediate guidance to address the neural oil painting problem, aiming to bridge the gap between the current iteration and the ultimate artistic vision, thereby enhancing the fidelity and effectiveness of the neural painting process.

In light of the aforementioned considerations, we adopt PaintTransformer~\cite{PaintTransformer} as our baseline and propose a new differential image-guided painter framework: the Differential Query Transformer (DQ-Transformer). The DQ-Transformer learns differential image features between the current canvas and the target image, focusing on the discrepancies between the images, thereby enabling more accurate stroke predictions.
In particular, we employ local encoders comprised of convolutional neural networks to learn three position-aware image features separately: the current canvas, the target image, and the differential image between these two. 
The differential image features are then transformed into query tokens, which are used as dynamic queries to the DQ-Transformer to decode the stroke parameters. 
The final painting result is obtained by rendering these decoded strokes onto the canvas. We first minimize the $L_1$ distance between the target image and the rendered image, as well as the $L_1$ distance between the predicted strokes and the ground-truth strokes.
Furthermore, we train the DQ-Transformer with a WGAN-based discriminator~\cite{Compositional_Neural_Painter,wgan}. 
The discriminator is utilized during training to enhance the precision of predicted strokes, by treating the rendered images as fake samples and striving to penalize the generation of erroneous strokes.
Compared with the baseline framework~\cite{PaintTransformer}, our DQ-Transformer retains its efficient inference advantage while innovatively introducing differential-guided dynamic queries. By explicitly focusing on image discrepancies through differential features, our method effectively eliminates duplicated stroke predictions and simultaneously captures subtle texture details.

The ``look, compare and draw'' painting process of our model is illustrated in Figure~\ref{fig:di}, where we present four intermediate stages of completing a real image with several strokes. It can be observed that our model evaluates the content of the differential image and introduces strokes precisely in areas exhibiting more significant disparities. This dynamic querying mechanism allows our model to prioritize areas that require refinement, progressively reducing visual differences and guiding the painting toward a highly detailed and structurally accurate final output.
Unlike existing stroke-based oil painting methods that often rely on static representations or fixed attention patterns, our approach is fundamentally \textbf{observation-first}: it continuously re-evaluates the evolving canvas in relation to the target, making each stroke placement both context-aware and purpose-driven. This design is conceptually simple and remarkably effective. 
To prove that the oil paintings produced by our method are of high quality, we compare them with other state-of-the-art stroke-based oil painting methods. Qualitative comparisons indicate that our method can generate images with more authentic oil painting textures while maintaining the fidelity of the original images. We have conducted a Mean Opinion Score (MOS) test and invited volunteers to evaluate the quality of oil paintings created by the above methods. The paintings of our method attained the highest preference ratings from the users. The primary contributions of our work are:

\begin{itemize}
\item  \textbf{Differential Image Analysis Integration:} 
We introduce a new painting pipeline that embeds differential image analysis within the neural oil painter framework. By focusing on the incremental changes wrought by successive brushstrokes, this simple and effective enhancement sharpens the attention to localized details, yielding a more intuitive and nuanced rendering process.
\item \textbf{Differential Query Transformer Architecture:} Inspired by the spirit of human artists, \ie, observing, comparing and drawing, we further introduce a Differential Query Transformer (DQ-Transformer) that explicitly leverages position-aware differential features as dynamic queries to guide stroke prediction.
\item \textbf{Superior Performance:} Both quantitative and qualitative experiments on three public datasets, \ie, Landscapes, FFHQ, and Wiki Art, affirm that the proposed method achieves better pixel-level and perception-level reconstruction, as well as higher user preference across various painting themes. Furthermore, the proposed method is stroke-efficient, \ie, it achieves competitive painting quality with fewer strokes. 
\end{itemize}

\section{Related Work}
\label{sec:related}
Stroke-based painting and pixel-wise painting represent two distinct paradigms in digital art creation. We first review related work on pixel-wise generation~\cite{PaintSeg,tvcg_chen2025human,zheng2026magic,zheng2026Progressive,zheng2024jointly}. To enhance robustness, DreamAnime~\cite{tvcg_xu2024dreamanime} disentangles anime style and identity into separate latent codes for independent text control. 3DArtmator~\cite{tvcg_zheng2024learning} and MVCGAN~\cite{zhang2023multi} incorporate 3D awareness through an interpretable stylization subspace and multi-view consistency, respectively. Huang~\etal~\cite{tvcg_huang2025creativesynth} propose a cross-art attention mechanism for style transfer, while DG-Net~\cite{zheng2019joint} disentangles style and content representations. For improved generation quality, Zhang~\etal~\cite{tvcg_zhang2023pose} introduce DPTN-TA, which uses dual-task correlation and a texture affinity loss for pose-guided person image synthesis and view synthesis. TextIR~\cite{tvcg_bai2025textir} leverages CLIP to align textual and visual features, achieving effective performance across multiple image restoration tasks. Despite their success, these pixel-based methods manipulate images holistically and do not reflect the stepwise, stroke-driven logic of human painting.

Unlike pixel-based generative models, automatic oil painting deploys brushstrokes as the fundamental unit of creation. Traditional stroke-based methods~\cite{haeberli1990paint,Litwinowicz_1997,collomosse2002painterly,hertzmann2001paint} rely on handcrafted rules to generate strokes. For example, Hertzmann~\etal~\cite{hertzmann1998painterly} apply multi-sized curved brush strokes to transform photographs into painterly renderings. Im2Oil~\cite{Im2Oil} combines adaptive sampling based on probability density maps to produce high-quality results. However, these rule-based approaches suffer from low search efficiency in large stroke spaces, leading to long runtimes.
Recently, deep learning based methods have gained increasing popularity, and various learning strategies have been explored to address stroke-based rendering. As noted by Hu~\etal~\cite{Compositional_Neural_Painter}, existing automatic oil painting methods based on deep neural networks can primarily be classified into three categories as follows:

\noindent\textbf{Optimization-based methods.} Optimization-based methods aim to determine the optimal stroke order to improve drawing efficiency. Fan~\etal~\cite{Fan_Animated} deconstruct brushstrokes in traditional Chinese ink paintings and introduce a natural evolution strategy to infer their best application sequence. To support stroke decomposition, Ashcroft~\etal~\cite{vectorstroke} propose a generative model for complex vector drawings and demonstrate its effectiveness on intricate anime line art. Stylized Neural Painting~\cite{StylizedNeuralPainting} treats stroke prediction as a parametric search process, mimicking a vector graphics renderer to adapt painting techniques to real images. Parameterized Brushstrokes~\cite{Parameterized} searches over parameterized stroke styles to complete a painting. Liu~\etal~\cite{tvcg_liu2023painterly} learn stroke style distributions and use semantic-aware placement to enhance artistic quality. Hertzmann~\etal~\cite{Hertzmann2024segmentation} leverage segmentation and dynamic attention maps to efficiently adjust stroke parameters.
These methods can be optimized jointly with neural style transfer but suffer from long optimization times for each image. 

\noindent\textbf{Neural network-based methods.} Neural network based methods directly use basic architectures to predict painting strokes. Early work employs Recurrent Neural Networks (RNNs)~\cite{rnn1} to decompose images into sequences, but relies on detailed manual annotations, limiting scalability. To overcome this, Frans~\etal~\cite{Frans_Sequence} apply self-supervised deep networks to learn the mapping from completed paintings to their brushstrokes. Paint Transformer~\cite{PaintTransformer} reformulates stroke prediction as a feed forward set generation task using a Transformer, enabling parallel stroke parameter prediction and efficient self supervised training without manual labels. Based on this work, Dong~\etal~\cite{dong2025domain} further study the efficient test-time adaptation. Similarly, Song~\etal~\cite{tvcg_song2024hairstyle} propose HairstyleNet, which combines parametric controllable strokes with neural rendering for high quality interactive hairstyle editing. Although these methods are annotation free and computationally efficient, their predicted strokes are often coarse and lack fine details near canvas boundaries.

\noindent\textbf{Reinforcement learning-based methods.} Reinforcement learning-based methods~\cite{Ganin_Kulkarni_Babuschkin_Eslami_Vinyals_2018,Zhou_Chen_Wang_Yang_Kim_Chen_Brandt_Terzopoulos_2018,SemaRL, Intelli-Paint, Intelligent-paint} aim to learn the textures and styles of real-world images to improve the painting quality. 
As a seminal effort, Huang~\etal~\cite{LearningtoPaint} employ a more complicated reinforcement learning model to paint complex real-world images with a watercolor brush. 
Moreover, Compositional Neural Painter~\cite{Compositional_Neural_Painter} incorporates object detection learning into the reinforcement learning model, dynamically segmenting and predicting stroke regions. Training a stable reinforcement learning agent is challenging due to the dynamic interactions among its components, as this process typically leads to instability. 

Although the aforementioned methods achieve satisfactory results in rendering paintings, they suffer from issues such as boundary inconsistencies and struggle with more intricate images. We address these limitations by introducing a DQ-Transformer architecture that leverages differentially derived image representations, augmented with positional information, to guide informed stroke prediction. Our model is both sensitive to position and capable of producing higher-quality renderings.

\section{Methodology}
\label{sec:method}

\begin{figure*}[t]
 \centering
 \includegraphics[width=0.90\linewidth]{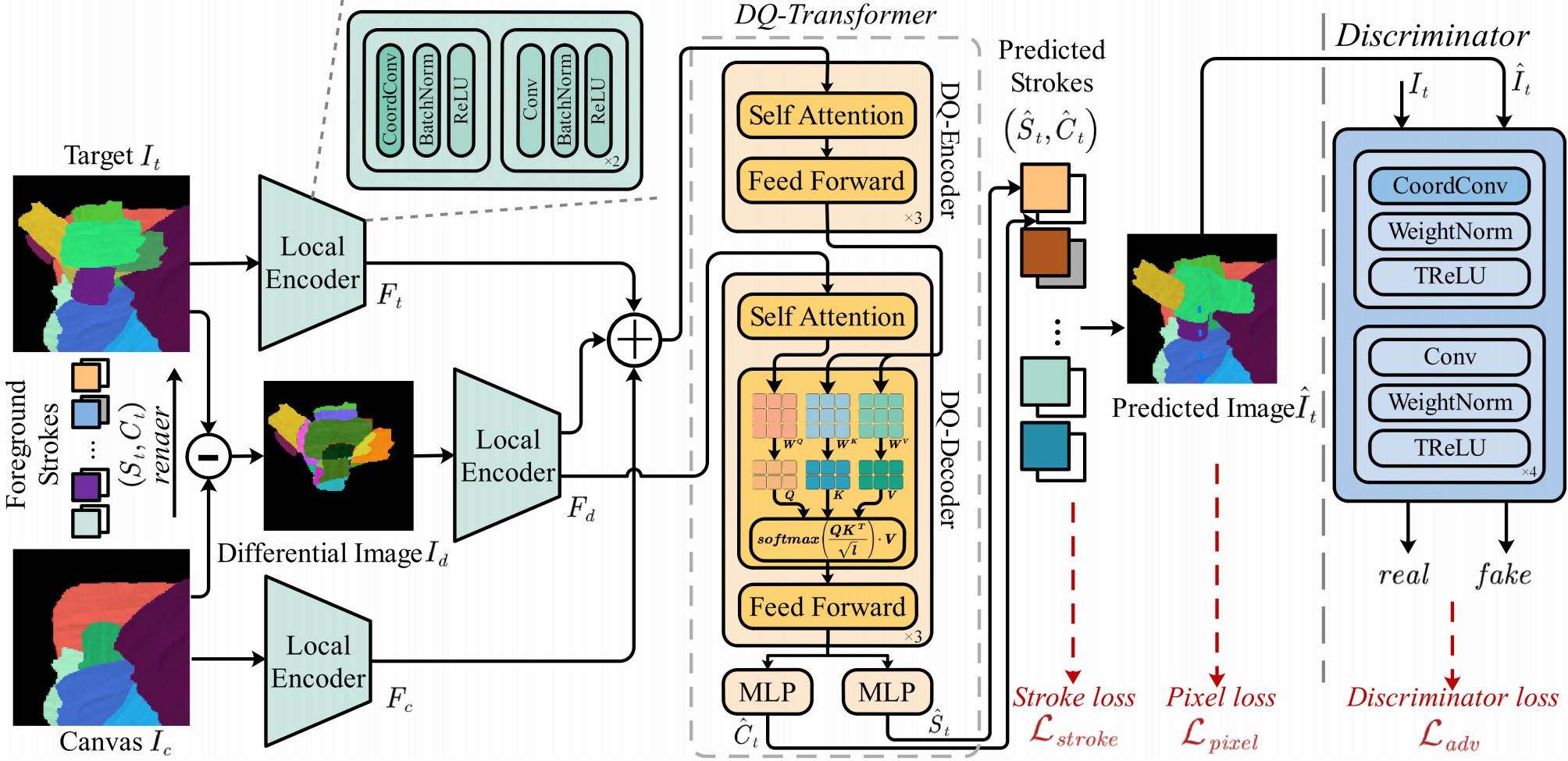}
 \caption{
 A brief overview of our painter framework. Given the canvas image $I_c$ and the target image $I_t$ generated by the renderer, we first obtain their differential image $I_d$ by simply subtracting one input from the other. Three local encoders comprised of convolutional neural networks are employed to extract image features $F_c,F_t,$ and $F_d$ with positional information. 
 DQ-Transformer has two components, \ie, the DQ-encoder and the DQ-decoder.
 These visual features $F_c,F_t$ and $F_d$, are concatenated and then fed to the DQ-encoder to obtain the fused feature $F_{kv}$.
 Next, we transform the differential image features $F_d$ into query tokens to query the key and value pairs generated by the fused feature $F_{kv}$. 
 Finally, the DQ-Transformer outputs a set of predicted strokes $\hat{S_t}$, each accompanied by its respective confidence $\hat{C_t}$. 
 The predicted image $\hat{I_t}$ is generated by rendering these strokes onto the canvas. The discriminator operates by treating the target images $I_t$ as real samples and the predicted images $\hat{I_t}$ as fake samples. 
 }
 \label{fig1}
\end{figure*}

\noindent\textbf{Overview.} Neural painting simplifies the painting task into predicting a sequence of brush strokes. In this section, we offer a comprehensive description of the training process for our painter framework, along with the inference process utilized for generating artworks. A brief overview of our painter framework is illustrated in Figure~\ref{fig1}. We utilize a self-supervised pipeline, originally introduced by \cite{PaintTransformer}, in which the current canvas and target images are constructed using randomly synthesized strokes, thereby eliminating the need for real images during the training process. Our objective is to guide the model to concentrate on the regions of discrepancy between the canvas and the target image, thereby predicting more accurate strokes to minimize these differences, without the necessity of considering the semantic information of the images.
Furthermore, we construct a differential image between the target image and the current canvas, which subsequently serves as the query tokens for our DQ-Transformer. The differential operation approximates how the human visual system processes image information, emphasizing the incremental effects resulting from consecutive brushstrokes. 

\subsection{Preliminaries}
\label{sec:sr}
\noindent\textbf{Stroke Renderer.} We adhere to the settings commonly employed in stroke-based painting methods \cite{StylizedNeuralPainting,PaintTransformer,Im2Oil,Compositional_Neural_Painter} for stroke rendering, adjusting the properties of real static brushstrokes,\ie, oil brushstrokes, to generate various stroke variants based on the specified parameters.
The stroke parameters are defined as $s=\left\{ x,y,h,w,\theta ,r,g,b \right\}$, where $\left( x,y \right) $ denotes the coordinates of the center point, $h$ represents the height, $w$ represents the width, $\theta$ denotes the rotation angle, and $\left( r,g,b \right)$ indicates the RGB color values of the stroke.
At each step $n$, the stroke renderer is employed to render the stroke parameters into a stroke image $R_n$ and a binary mask $M_n$, where $M_n$ is a single-channel alpha map of $R_n$. These stroke images are then sequentially added to the current canvas, potentially covering any previous strokes if they exist. The iterative rendering process can be formulated as:
\begin{equation}
I_{n}=R_n\odot c_n M_n + I_{n-1}\odot \left( 1-c_n M_n \right),
\end{equation}
where $c_n$ is the confidence of the stroke, indicating whether the stroke is valid. $\odot$ is the element-wise multiplication, while $I_{n-1}$ is the previous painting result. The entire rendering process is based on differentiable linear transformations and does not contain  any trainable parameters.

\noindent\textbf{Canvas Construction.} In each training iteration, we first randomly sample two strokes sets: a background strokes set $S_b$ to generate the canvas $I_c$, and a foreground strokes set $S_t$ to create the target image $I_t$ based on $I_c$. 
Background strokes are rendered onto an empty canvas to establish the current canvas $I_c$. Subsequently, the foreground strokes are superimposed onto the current canvas to produce the target image $I_t$. Notably, the background strokes are coarser in granularity than the foreground strokes. This construction methodology mirrors the human artistic process, which evolves from broad outlines to detailed refinements.

\subsection{Painter Framework}
The painter framework aims to reconstruct the target image $I_t$  using a sequence of predicted strokes. 
Given the current canvas $I_c\in \mathbb{R}^{3\times P\times P}$ and the target image $I_t\in \mathbb{R}^{3\times P\times P}$, where $P$ is the pre-defined patch size that acts as the basic unit for subsequent painting. Then the differential image is obtained by performing a pixel-wise subtraction: $I_d = I_t - I_c$. Our painter framework takes $I_c,I_t,$ and $I_d$ as input and predicts a stroke set $\hat{S_t}$. The predicted image is generated by rendering these strokes onto the canvas.

\noindent\textbf{Local Encoder.} As shown in Figure~\ref{fig1}, the painter framework first employs separate local encoders, comprised of convolutional neural networks, to individually extract their feature maps, denoted as $F_c,F_t,F_d\in \mathbb{R}^{3\times {\frac{P}{4}\times \frac{P}{4}}}$. It is worth noting that traditional convolutional layers lack explicit positional encoding, and stacking them directly can lead to the loss of coordinate information. To address this issue, we substitute traditional convolutional layers with Coordinate Convolution (CoordConv)~\cite{CoordConv}, implementing it in the first layer of the convolutional network. CoordConv introduces additional channels to the input feature map, representing the X-Y coordinates of each feature pixel, thereby enabling the convolutional learning process to have a degree of awareness about the spatial positions. Then, $F_c,F_t,$ and $F_d$, endowed with positional encoding, are concatenated and flattened as the input of DQ-Transformer. 

\noindent\textbf{DQ-Transformer.} DQ-Transformer consists of two main parts: a DQ-Encoder and a DQ-Decoder. The DQ-Encoder block consists of a self-attention layer and a feed-forward layer, and it learns a mapping from the concatenated features \{$F_c$, $F_t$, $F_d$\} to produce the fused features $F_{kv}$.
The DQ-Decoder block comprises a self-attention layer, a cross-attention layer, and a feed-forward layer. In the DQ-Decoder, the differential image features $F_d$ are transformed into query tokens. This transformation helps the model focus on local changes introduced by incremental strokes. The DQ-Decoder then considers the correspondences between the differential query tokens $F_d$ and the fused features $F_{kv}$ output by the DQ-encoder. 
The self-attention layer learns the relative attention and interactions among the various elements of differential query tokens. 
The cross-attention layer implements $CrossAttention\left( Q;K;V \right) =softmax\left( {\frac{QK^T}{\sqrt{l}}} \right)~\cdot~V$, and $l$ is the output dimension of key and query features, while
\begin{equation}
Q=W^QF_d,K=W^KF_{kv},V=W^VF_{kv},
\end{equation}
where $W^Q$, $W^K,$ and $W^V$ are learnable weights that project $F_d$ to query, and map $F_{kv}$ to key and value, respectively.
Finally, the differential query tokens are fed through two MLPs to predict stroke parameters $\hat{S_t}=\left\{ \hat{s}_i \right\} _{i=1}^{N}$ and their corresponding confidences $\hat{C_t}=\left\{ \hat{c}_i \right\} _{i=1}^{N}$ respectively. 
During the inference phase, we determine whether the predicted stroke is valid based on the sign of confidence $\hat{c}_i$.
If $\hat{c}_i\geqslant 0$, we draw this stroke, otherwise, we skip it. We draw all predicted valid strokes onto the canvas, yielding the final painting $\hat{I_t}$.

\subsection{Training Objective} 

\noindent\textbf{Pixel Loss.}
The most direct goal of neural painting is to reconstruct the target image. Therefore, similar to \cite{PaintTransformer,Intelli-Paint}, we minimize the $L_1$ distance between the predicted image $ \hat{I_t}$ and the target image $I_t$ as:
\begin{equation} 
\mathcal{L}_{pixel}=\lambda_p\left\| I_t  -\hat{I_t} \right\| _1,
\end{equation}
where $\lambda_p$ is a weight term.

\noindent\textbf{Stroke Loss.}
Given that the target image is rendered from the canvas image using the set of foreground strokes, we can constrain the difference between the ground-truth and the prediction at the stroke level. We follow the stroke loss~\cite{PaintTransformer} on the re-matched strokes as:
\begin{equation}
\begin{split}
\mathcal{D}_{match} = \frac{1}{|S_t|}  \sum_{u=1}^{|S_t|} & \left( 
    c_{u} \left( \mathcal{D}_{L_1}^{u} + \lambda_{W} \mathcal{D}_{W}^{u} \right) + \mathcal{D}_{bce}^{u} \right) ,
\end{split}
\end{equation}
where $u$ and $\hat{u}$ represent the target strokes and predicted strokes respectively. 
$\mathcal{D}_{L_1}^{u}, \mathcal{D}_{W}^{u}$, and $\mathcal{D}_{bce}^{u}$ represent the pixel loss, rotation loss, and classification loss of the stroke set, respectively, as proposed by~\cite{PaintTransformer}.
$\lambda_{W}$ is a weight term, and $|S_t|$ is the number of strokes. 

Further, to encourage the model to reconstruct the target using the minimum number of valid strokes, we impose an additional regularization on the confidence $\hat{C_t}$ of the predicted strokes. 
We derive the stroke loss:
\begin{equation}
\mathcal{L} _{stroke}=\mathcal{D} _{match}+\lambda _{c}{\frac{1}{|S_t|}}\sum_{u=1}^{|S_t|}{\left\| \hat{c}_u \right\| _1},
\label{eq:L1}
\end{equation}
where $\lambda _c$ is a weight term for the confidence regularization.

\noindent\textbf{Adversarial Loss.}
Treating our painting network as a generator, we design a simple discriminator, which regards the generated images as fake samples, encouraging the model to predict strokes that make the painting closer to the target image. 
As shown in Figure~\ref{fig1}, the discriminator consists of five blocks. 
In the first block, we replace the Conv layer with a CoordConv layer.
The training process employs a WGAN-GP loss~\cite{Compositional_Neural_Painter} as:
\begin{equation}
\mathcal{L} _{adv}=Dis\left( \hat{I}_t \right) -Dis\left( I_t \right) +\lambda _{dis}\left( \left\| \nabla _{\tilde{I}_t}Dis\left( \tilde{I}_t \right) \right\| _2-1 \right) ^2,
\label{eq:adv}
\end{equation}
where $Dis\left( \cdot \right)$ represents the discriminator score for a given sample. $\tilde{I}_t$ is a linear interpolation between real samples $I_t$ and fake samples $\hat{I_t}$. $ \left\| \nabla _{\tilde{I}_t}Dis\left( \tilde{I}_t \right) \right\|_2$ is the $L2$ norm of the gradient of the discriminator on the interpolation point. $\lambda _{dis}$ is the hyperparameter for the gradient penalty.

\noindent\textbf{Overall loss.}
Finally, our network is optimized by the pixel loss, the stroke loss, and the adversarial loss as:
\begin{equation}
\mathcal{L} _{total}=\mathcal{L} _{pixel}+\mathcal{L} _{stroke}+\gamma \mathcal{L} _{adv},
\end{equation}
where $\gamma =\frac{\left\| \mathcal{L} _{pixel} \right\|}{\left\| \mathcal{L} _{adv} \right\|}
$ is an adaptive balancing factor~\cite{Compositional_Neural_Painter}.

\subsection{Painting Inference} 
\begin{figure}[!t]
\centering
    \includegraphics[width=0.48\textwidth]{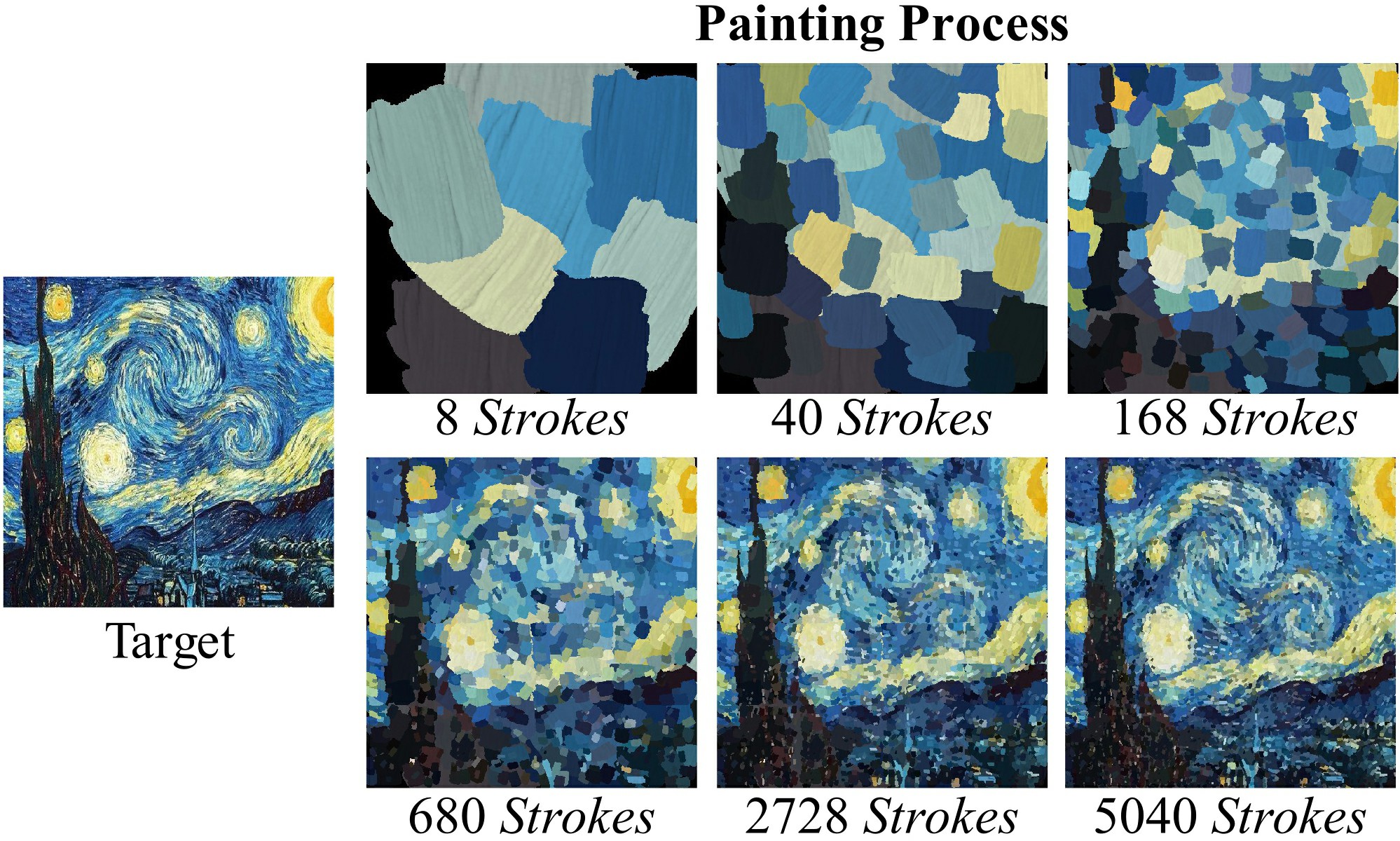}
    \caption{Our painting progress following a coarse-to-fine manner.}
    \label{fig:pp}
\end{figure}

Following the painting strategies of \cite{StylizedNeuralPainting,PaintTransformer}, our model generates paintings in a progressive manner, starting from a coarse sketch and gradually refining details across multiple scales. It is worthy noting that the stroke number of our method is not fixed. Because our network also predicts ``skip'' when the current painting area is already satisfactory. Our coarse-to-fine painting process is illustrated in Figure~\ref{fig:pp}. Moreover, our method produces seamless results without visible patch seams. This is enabled by two design choices: (1) the use of spatial positional embeddings that preserve location awareness even near patch edges; (2) our differential-query mechanism, which conditions stroke prediction on the residual error map across the full canvas context. As a result, strokes near boundaries are not suppressed, and the final composition remains visually coherent.

\section{Experiment}
\subsection{Implementation Details}
\noindent\textbf{Datasets.} 
Our model is trained exclusively using synthesized stroke images, without relying on any real-world datasets. We conduct evaluation on three distinct datasets: Landscapes~\cite{cartoongan}, FFHQ~\cite{ffhq}, and Wiki Art~\cite{WikiArt}. The Landscapes dataset comprises the natural landscape images sourced from the Flickr website. FFHQ is a high-quality face image dataset that covers a variety of ages, genders, races, and expressions. WikiArt is a compilation comprising a large number of artistic pieces with diverse styles, each piece created through genuine human painting. 
For each dataset, we randomly select 100  images as test samples.

\noindent\textbf{Settings.} 
We set patch size $P$ as 32 and the maximum number of brushstrokes $|S_t|$ in one patch as 8. 
During training, parameters for target strokes are randomly generated from a uniform distribution. 
We sequentially render these strokes, and if a stroke covers more than 75\% of the area of the preceding stroke, its confidence is set to 0 to ensure that the rendered strokes do not overly overlap. 
We follow existing works~\cite{PaintTransformer} to set hyper-parameters $\lambda _p=8$, and $\lambda _W=10$. For the adversarial loss weight, we follow~\cite{Compositional_Neural_Painter} and set $\lambda _{dis}=10$. 
We have conducted experiments to determine the appropriate weight in Eq.~\ref{eq:L1} and ultimately set $\lambda _c=0.1$  as default.
We use the AdamW optimizer~\cite{adamw} with an initial learning rate of $1 \times 10^{-4}$ and set weight decay to $1 \times 10^{-2}$. The model is trained for 100,000 iterations using a batch size of 64. The first 50,000 iterations are dedicated to pre-training the painting network without the adversarial loss. This strategy helps to avoid mode collapse, ensuring that the generator can faithfully reconstruct the target images.

\begin{figure*}[htp]
\centering
    \includegraphics[width=0.9\textwidth]{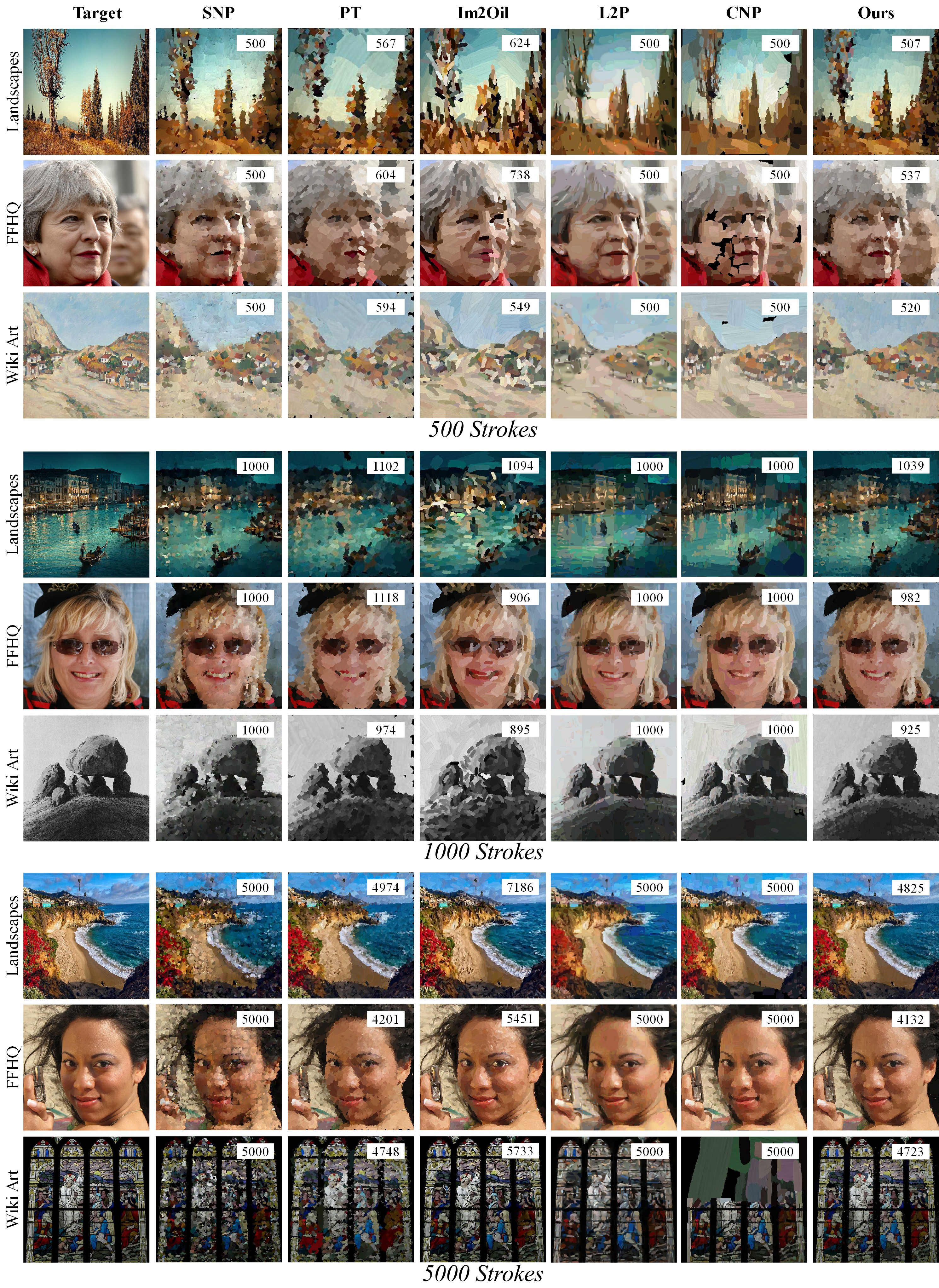}
    \caption{Qualitative comparison between our model and state-of-the-art neural painting methods on unseen real-world datasets at different levels of stroke counts. The actual number of strokes used in the painting is annotated in the top right corner of the image. Our method leverages the difference image as a dynamic query for each painting step. This observation-first approach enables our model to achieve superior visual quality with relatively fewer strokes, effectively reproducing complex details with high fidelity. Please zoom in to obtain a more detailed view.}
    \label{fig2}
\end{figure*}

\begin{table*}[t]
  \centering
  \caption{Quantitative comparison with competitive methods under pixel-level and perception-level reconstruction on unseen real-world datasets at different levels of stroke counts. Lower values indicate better reconstruction. Bold indicates best. Benefiting from the observation-first mechanism, our model adapts to varying stroke budgets while preserving fine-grained details and global structure, consistently achieving strong performance across a wide range of stroke counts. This shows its robustness and efficiency in high-fidelity neural painting under resource constraints.}
  \resizebox{0.97\linewidth}{!}{
    \begin{tabular}{c|l||cc||cc||cc||cc}
    \toprule
          &       & \multicolumn{2}{c||}{Landscape} & \multicolumn{2}{c||}{FFHQ} & \multicolumn{2}{c||}{Wiki Art } & \multicolumn{2}{c}{Average} \\
    Stroke & \multicolumn{1}{c||}{Method} & $\mathcal{L}_{pixel}\downarrow$ & $\mathcal{L}_{pcpt}\downarrow$  & $\mathcal{L}_{pixel}\downarrow$ & $\mathcal{L}_{pcpt}\downarrow$  & $\mathcal{L}_{pixel}\downarrow$ & $\mathcal{L}_{pcpt}\downarrow$  & $\mathcal{L}_{pixel}\downarrow$ & $\mathcal{L}_{pcpt}\downarrow$ \\
    \midrule
          & Stylized Neural Painting & 0.068  & 0.941  & 0.057  & 1.047  & 0.064  & 0.998  & 0.063  & 0.995  \\
          & Paint Transformer & 0.080  & 0.851  & 0.067  & 1.052  & 0.072  & 0.934  & 0.073  & 0.946  \\
    500   & Im2Oil & 0.096  & 0.992  & 0.077  & 1.071  & 0.089  & 1.036  & 0.087  & 1.033  \\
          & Learning To Paint & 0.065  & 0.793  & \textbf{0.050}  & \textbf{0.850}  & 0.062  & 0.833  & 0.059  & 0.825  \\
          & Compositional Neural Painter & 0.069  & 0.886  & 0.053  & 0.996  & 0.062  & 0.907  & 0.062  & 0.930  \\
          & \textbf{Ours} & \textbf{0.063} & \textbf{0.751} & 0.051 & 0.881 & \textbf{0.058} & \textbf{0.812} & \textbf{0.057} & \textbf{0.815} \\
    \midrule
          & Stylized Neural Painting & 0.072  & 0.921  & 0.060  & 1.012  & 0.067  & 0.974  & 0.066  & 0.969  \\
          & Paint Transformer & 0.079  & 0.843  & 0.064  & 1.045  & 0.069  & 0.913  & 0.071  & 0.934  \\
    1000  & Im2Oil & 0.094  & 0.983  & 0.071  & 1.040  & 0.087  & 1.022  & 0.084  & 1.015  \\
          & Learning To Paint & 0.063  & 0.805  & \textbf{0.046}  & 0.833  & 0.057  & 0.829  & 0.055  & 0.822  \\
          & Compositional Neural Painter & 0.063  & 0.848  & 0.048  & 0.946  & 0.056  & 0.864  & 0.056  & 0.886  \\
          & \textbf{Ours} & \textbf{0.062} & \textbf{0.751} & 0.047 & \textbf{0.830} & \textbf{0.056} & \textbf{0.789} & \textbf{0.055} & \textbf{0.790} \\
    \midrule
          & Stylized Neural Painting & 0.068  & 0.939  & 0.057  & 1.044  & 0.064  & 0.996  & 0.063  & 0.993  \\
          & Paint Transformer & 0.070  & 0.807  & 0.056  & 0.934  & 0.061  & 0.841  & 0.062  & 0.861  \\
    5000  & Im2Oil & 0.064  & 0.720  & 0.042  & 0.742  & 0.052  & 0.718  & 0.053  & 0.727  \\
          & Learning To Paint & 0.055  & 0.718  & \textbf{0.032}  & 0.697  & 0.047  & 0.705  & \textbf{0.045}  & 0.707  \\
          & Compositional Neural Painter & 0.056  & 0.732  & 0.037 & 0.772  & 0.046  & 0.715  & 0.046  & 0.740  \\
          & \textbf{Ours} & \textbf{0.054} & \textbf{0.579} & 0.039  & \textbf{0.631} & \textbf{0.045} & \textbf{0.593} & 0.046 & \textbf{0.601} \\
    \bottomrule
    \end{tabular}%
    }
  \label{tab:t1}%
\end{table*}%

\subsection{Comparison with State-of-the-Art Methods}
\label{sec:csoat}

\noindent\textbf{Quantitative Comparison.} 
We conduct a quantitative comparison between our method and four state-of-the-art oil painting methods: Stylized Neural Painting~\cite{StylizedNeuralPainting} (an optimization-based model), Paint Transformer~\cite{PaintTransformer} (a neural network-based model), Im2Oil~\cite{Im2Oil} (a traditional search-based model), Learning to Paint~\cite{LearningtoPaint} (a reinforcement learning-based model) and Compositional Neural Painter~\cite{Compositional_Neural_Painter} (a reinforcement learning-based model). Since the main objective of neural painting is to recreate original images, we directly use the pixel loss $\mathcal{L}_{pixel}$ and the perceptual loss $\mathcal{L}_{pcpt}$~\cite{Perceptual} as evaluation metrics. 
$\mathcal{L}_{pixel}$ calculates the mean $L_1$ distance between the rendered images and the target images at the pixel level. $\mathcal{L}_{pcpt}$ is a perceptual metric based on neural network features, which measures the similarity between a target image and a generated image by comparing their differences in high-level feature maps. Lower values of $\mathcal{L}_{pixel}$ and $\mathcal{L}_{pcpt}$ both indicate a better image reconstruction quality. 
All painting results are produced at a resolution of $512 \times 512 $ pixels. Among the five methods we compare, Stylized Neural Painting, Learning to Paint, and Compositional Neural Painter can set the exact number of strokes. Paint Transformer and Im2Oil can only roughly control the number of strokes by adjusting the setting parameters. 
For a fair comparison, we conduct experiments at 500, 1,000, and 5,000 strokes respectively.

Table~\ref{tab:t1} shows our results on various datasets at different levels of stroke counts.
It is intriguing to observe that all methods exhibit loss fluctuations across different datasets, indicating a substantial influence of image content complexity on the painting results. 
For example, our paintings achieve a lower pixel loss and a higher perceptual loss on the FFHQ dataset compared to the Landscapes and Wiki Art datasets.
This difference can be attributed to the nature of the images in each dataset. Although plein-air paintings from the Landscapes dataset exhibit complex compositions, they possess less high-level semantic information compared to the high-definition facial images in the FFHQ dataset. Consequently, the plein-air paintings experience higher pixel loss but lower perceptual loss. This also illustrates the necessity of incorporating both pixel and perceptual loss as evaluation metrics, as they capture different aspects of the painting quality.

As shown in Table~\ref{tab:t1}, our approach achieves the best overall balance: at 500 strokes, we obtain the lowest average perceptual loss (0.815) and competitive pixel loss (0.057); at 5000 strokes, we further reduce perceptual loss to 0.601, significantly outperforming all baselines. Notably, Learning to Paint achieves the lowest pixel loss on FFHQ, which can be attributed to its training on additional human face datasets (\eg, CelebA~\cite{celeb}) as reported in the original work. In contrast, our method is trained solely on random strokes without any domain-specific images, yet still achieves competitive pixel accuracy (0.039 at 5000 strokes) while significantly outperforming Learning to Paint in perceptual quality (0.631 vs. 0.697 on FFHQ).
While methods such as Stylized Neural Painting, Paint Transformer, Im2Oil, and Compositional Neural Painter achieve competitive results in certain settings, their overall performance remains inferior to ours.  
The quantitative results highlight the robustness and effectiveness of our approach in reconstructing high-quality images under increasingly complex stroke configurations.

\noindent\textbf{Qualitative Comparison.} 
Figure~\ref{fig2} presents a comprehensive qualitative comparison across three diverse image categories and three stroke budgets (500, 1000, 5000). Stylized Neural Painting produces blocky results with visible grid artifacts, especially at high stroke counts, and yields blurred facial details on FFHQ. Paint Transformer generates coarse strokes that miss fine structures, leading to poor edge definition across all datasets. Im2Oil over-samples strokes in textured regions, such as sand or hair, causing cluttered and disordered outputs due to its density-based sampling strategy. Learning to Paint achieves low pixel loss on faces by leveraging extra face-specific training data, but its renderings appear over-smoothed and airbrushed, lacking authentic brushstroke expressiveness. Compositional Neural Painter, relying on object priors, often leaves blank regions or misaligns strokes on novel or complex scenes like WikiArt, indicating limited generalization. In contrast, our method accurately reconstructs image content while preserving vivid and coherent brushwork, requires no domain-specific image data, and consistently delivers superior visual quality across diverse image types and stroke budgets.

\begin{figure*}[tp]
\centering
    \includegraphics[width=0.9\textwidth]{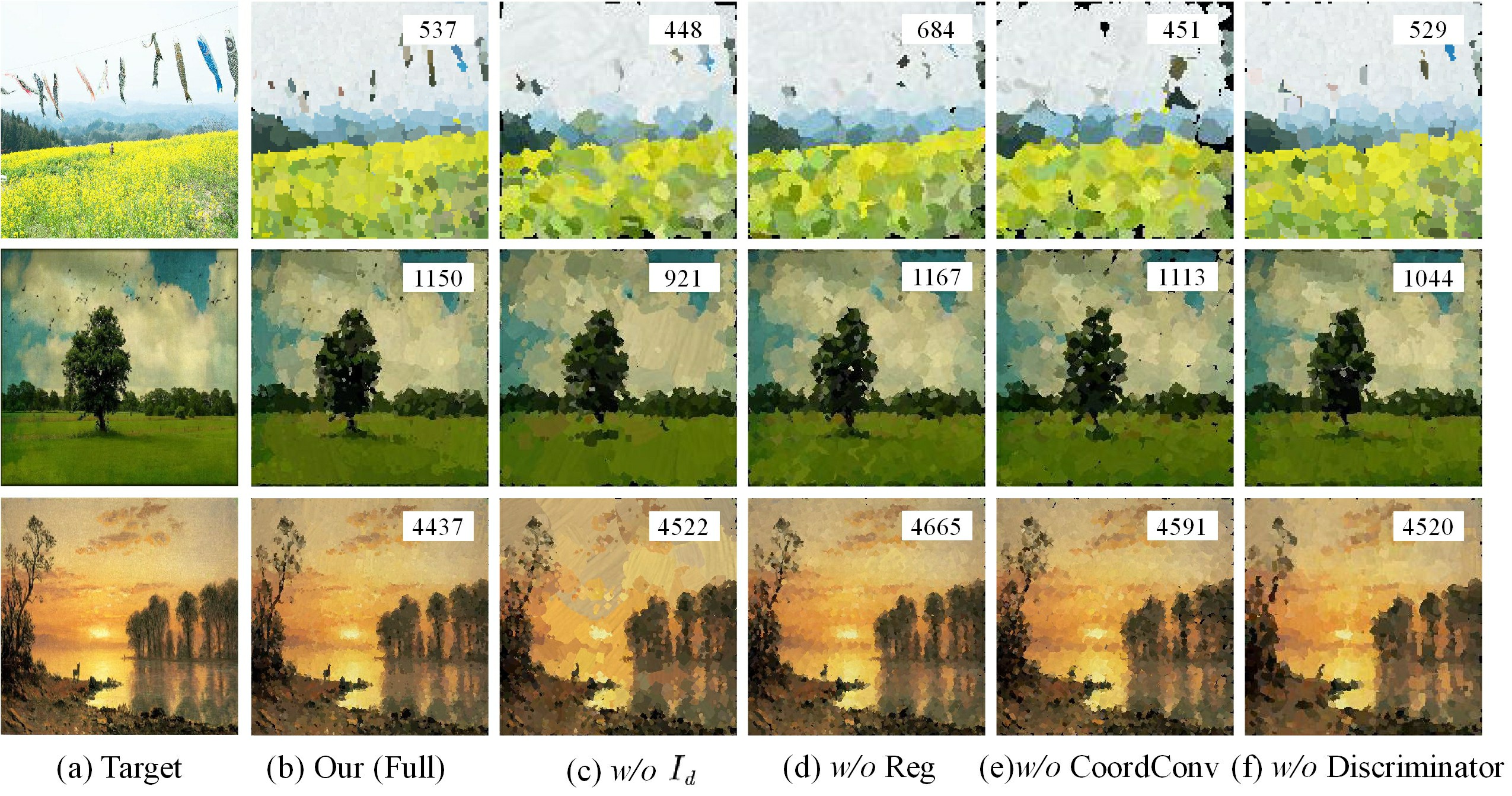}
    \vspace{-.1in}
    \caption{Ablation study on the primary components of our framework at different stroke counts. The actual number of brushstrokes used in the painting is annotated in the top right corner of the image. Please zoom in to obtain a more detailed view.}
    \label{fig:ab}
\end{figure*}

\begin{figure}[!t]
\centering
    \includegraphics[width=0.48\textwidth]{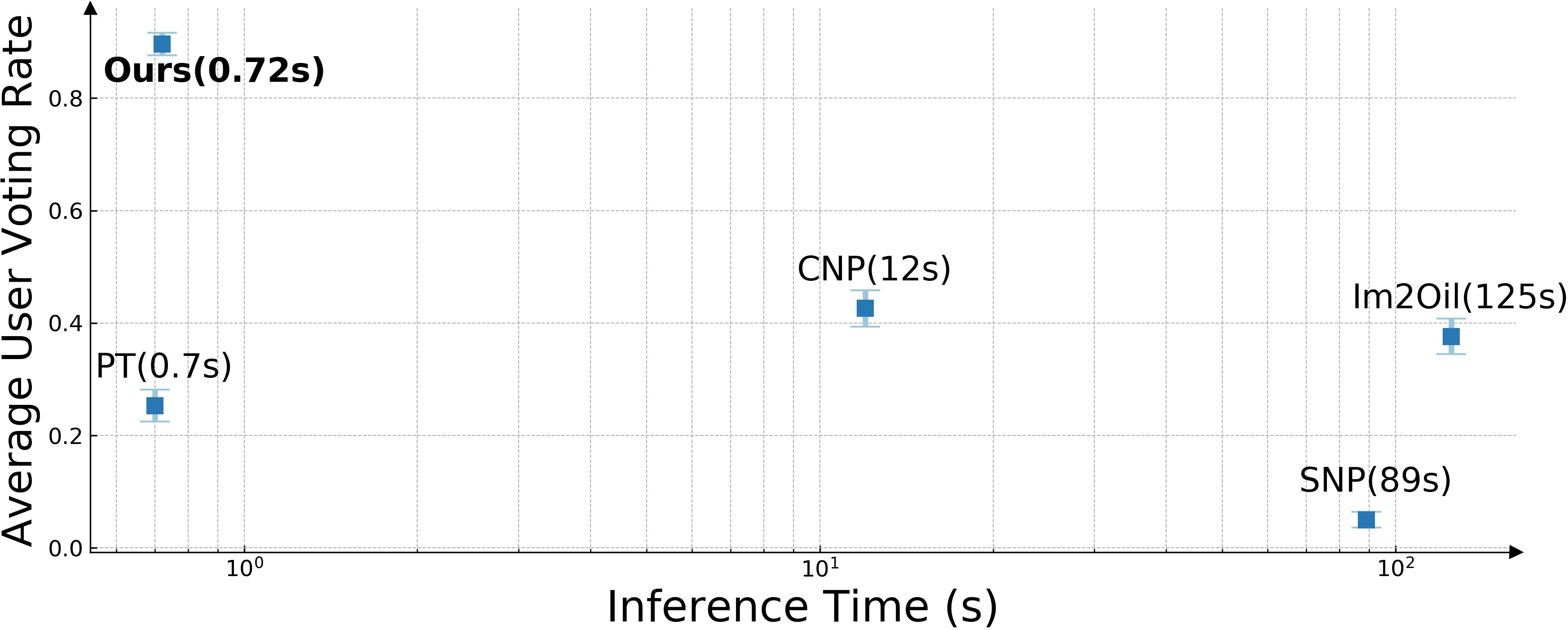}
    \vspace{-.1in}
    \caption{The average user voting rate and inference time for each method. Methods positioned closer to the upper left corner are characterized by higher user votes and faster inference speeds. Our approach surpasses the comparison methods in preference score by a clear margin and also offers faster inference speed.}
    \label{fig:mos}
\end{figure}

\noindent\textbf{User Study.} 
To further evaluate the painting quality of our model, we conducted a Mean Opinion Score (MOS) study~\cite{Im2Oil} to assess user preferences for automatic oil painting methods. We recruit a total of 30 graduate students from diverse disciplines across our university to participate in the MOS test.  
We launch a questionnaire website through Gradio~\cite{gradio}. Each questionnaire involves the random selection of 30 image sets, wherein each set comprises one target image alongside five corresponding oil paintings. The identities of these five oil paintings are anonymized within each set, and their presentation sequence is randomized to mitigate order effects. Participants are instructed to evaluate each set of oil paintings and identify the two works they deem to exhibit superior quality. By limiting participants to selecting their top-2 choices, we aim to focus on the most outstanding results while avoiding the potential ambiguity and difficulty of ranking lower-quality paintings. The average user voting rate of each method is shown on the vertical axis of Figure~\ref{fig:mos}. Collectively, the data indicate a pronounced user preference for our proposed oil painting method relative to alternative approaches. Although Compositional Neural Painter and Im2Oil show commendable painting quality, their performance is inconsistent across different images, leading to slightly lower votes. Stylized Neural Painting and Paint Transformer have limitations in detail rendering, which negatively impacts their overall voting.

\noindent\textbf{Efficiency Analysis.} The training and inference times of all methods, measured on a single NVIDIA RTX 3090 Ti GPU using their official implementations and default settings, are summarized in Table~\ref{tab:te}. 
Our method achieves a training time of only 10 hours, substantially outpacing reinforcement learning–based approaches such as Learning to Paint (50 hours) and Compositional Neural Painter (90 hours), which rely on costly policy optimization and extensive environment exploration. It further surpasses Stylized Neural Painting in efficiency and matches the training speed of Paint Transformer. At inference time, our approach renders each image in just 0.72 seconds, on par with Paint Transformer and orders of magnitude faster than Stylized Neural Painting, Im2Oil, and Compositional Neural Painter. Notably, this high computational efficiency is attained without compromising visual fidelity.

\begin{table*}[!t]
  \centering
  \caption{Quantitative Effect of Primary Components at different levels of stroke counts. \textit{w/o} $I_d$ denotes that we do not use the differential image, while \textit{w/o} Reg ($\lambda_c = 0$) means the model without confidence regularization in Eq.~\ref{eq:L1}, \textit{w/o} CoordConv represents we solely employ conventional convolutional layers to extract image features, \textit{w/o} Discriminator denotes that we train the model without the discriminator.}
  \resizebox{0.97\linewidth}{!}{
    \begin{tabular}{c|l||cc||cc||cc||cc}
    \toprule
          &      & \multicolumn{2}{c||}{Landscape} & \multicolumn{2}{c||}{FFHQ} & \multicolumn{2}{c||}{Wiki Art } & \multicolumn{2}{c}{Average} \\
    Stroke & \multicolumn{1}{c||}{Method} & $\mathcal{L}_{pixel}\downarrow$ & $\mathcal{L}_{pcpt}\downarrow$  & $\mathcal{L}_{pixel}\downarrow$ & $\mathcal{L}_{pcpt}\downarrow$  & $\mathcal{L}_{pixel}\downarrow$ & $\mathcal{L}_{pcpt}\downarrow$  & $\mathcal{L}_{pixel}\downarrow$ & $\mathcal{L}_{pcpt}\downarrow$ \\
    \midrule
          & \textit{w/o} $I_d$ & 0.095  & 0.883  & 0.088  & 1.079  & 0.090  & 0.935  & 0.091  & 0.966  \\
          & \textit{w/o} Reg ($\lambda_c = 0$) & 0.085  & 0.849  & 0.077  & 1.048  & 0.081  & 0.907  & 0.081  & 0.935  \\
    500   & \textit{w/o} CoordConv & 0.103  & 0.935  & 0.104  & 1.145  & 0.101  & 0.994  & 0.103  & 1.025  \\
          & \textit{w/o} Discriminator & 0.070  & 0.920  & 0.062  & 1.037  & 0.065  & 0.975  & 0.066  & 0.977  \\
          & \textbf{Ours (Full)} & \textbf{0.063} & \textbf{0.751} & \textbf{0.051} & \textbf{0.881} & \textbf{0.058} & \textbf{0.812} & \textbf{0.057} & \textbf{0.815} \\
    \midrule
          & \textit{w/o} $I_d$ & 0.084  & 0.858  & 0.071  & 1.027  & 0.075  & 0.899  & 0.077  & 0.928  \\
          & \textit{w/o} Reg ($\lambda_c = 0$) & 0.072  & 0.794  & 0.058  & 0.949  & 0.065  & 0.837  & 0.065  & 0.860  \\
    1000  & \textit{w/o} CoordConv & 0.086  & 0.911  & 0.076  & 1.081  & 0.082  & 0.959  & 0.081  & 0.984  \\
          & \textit{w/o} Discriminator & 0.068  & 0.821  & 0.055  & 0.916  & 0.061  & 0.867  & 0.061  & 0.868  \\
          & \textbf{Ours (Full)} & \textbf{0.062} & \textbf{0.751} & \textbf{0.047} & \textbf{0.830} & \textbf{0.056} & \textbf{0.789} & \textbf{0.055} & \textbf{0.790} \\
    \midrule
          & \textit{w/o} $I_d$ & 0.078  & 0.833  & 0.064  & 0.975  & 0.066  & 0.868  & 0.069  & 0.892  \\
          & \textit{w/o} Reg ($\lambda_c = 0$) & 0.064  & \textbf{0.476}  & 0.048  & 0.791  & 0.055  & 0.736  & 0.056  & 0.668  \\
    5000  & \textit{w/o} CoordConv & 0.075  & 0.854  & 0.059  & 0.976  & 0.067  & 0.899  & 0.067  & 0.910  \\
          & \textit{w/o} Discriminator & 0.059  & 0.735  & 0.047  & 0.713  & 0.051  & 0.770  & 0.052  & 0.739  \\
          & \textbf{Ours (Full)} & \textbf{0.054} & 0.579 & \textbf{0.039} & \textbf{0.631} & \textbf{0.045} & \textbf{0.593} & \textbf{0.046} & \textbf{0.601} \\
    \bottomrule
    \end{tabular}%
    }
  \label{tab:t1_1}%
\end{table*}%

\subsection{Ablation Studies}
\noindent\textbf{Quantitative Effect of Primary Components.} To validate the effectiveness and robustness of each component in our framework, we conduct an extensive ablation study across three representative stroke budgets: 500, 1000, and 5000 strokes. We train four ablated models: one variant without the differential image; one variant without the confidence regularization in Eq.~\ref{eq:L1}; one variant without CoordConv layers; and one variant without the WGAN-based discriminator. As shown in Table~\ref{tab:t1_1}, removing the differential image leads to the most significant degradation, especially under low stroke budgets (\eg, +0.082 in $L_{pixel}$ at 500 strokes), highlighting that error-driven dynamic queries are essential for guiding efficient stroke placement. This confirms that our formulation enables the model to focus directly on reconstruction residuals, improving sample efficiency. The benefit of confidence regularization becomes increasingly evident as the stroke budget grows: its absence leads to higher perceptual loss at 5,000 strokes on both FFHQ and WikiArt. Similarly, discarding the adversarial loss degrades performance on complex WikiArt scenes, where fine textural details are critical. Crucially, the full model maintains a consistent advantage over all ablated versions across all stroke budgets, underscoring the complementary roles of each component in achieving both high fidelity and rendering efficiency.

\noindent\textbf{Qualitative Effect of Primary Components.} The qualitative results are shown in Figure~\ref{fig:ab}. Without the differential image, the model suffers from redundant strokes and poor refinement, as seen in the over-painted grass regions. Removing confidence regularization leads to noisy and unstable stroke generation, particularly evident in fine details like tree edges. The absence of CoordConv degrades spatial coherence, resulting in blurred boundaries and distorted structures. Finally, eliminating the discriminator causes a loss of artistic style and perceptual realism, producing less expressive paintings. 
Upon zooming into the detailed sections of the image, the painting produced by the full model appears smoother.

\begin{table}[t]
  \centering
  \caption{Comparison of training and inference time across different painting methods.}
    \begin{tabular}{c|cccccc}
    \toprule
    Method & SNP   & PT    & Im2Oil & L2P   & CNP   & Ours \\
    \midrule
    Training (\emph{hours}) & 11    & 4     & -     & 50    & 90    & 10 \\
    \midrule
    Inference (\emph{seconds}) & 89    & 0.70   & 125   & 3     & 12    & 0.72  \\
    \bottomrule
    \end{tabular}
  \label{tab:te}
\end{table}

\begin{table}[t]
  \centering
  \caption{Ablation study on the weight $\lambda_c$. We set $\lambda _c=0.1$  as the default value.}
    \begin{tabular}{cccccccc}
    \toprule
    $\lambda_c$     & 0.05  & 0.1   & 0.2   & 0.5   & 1     & 5     & 10 \\
    \midrule
    $\mathcal{L}_{pixel}\downarrow$ & 0.048 & \textbf{0.046} & \textbf{0.046} & 0.050  & 0.050  & 0.055 & 0.058  \\
    $\mathcal{L}_{pcpt}\downarrow$ & 0.668 & \textbf{0.607} & 0.614 & 0.686 & 0.685 & 0.786 & 0.791  \\
    \bottomrule
    \end{tabular}%
  \label{tab:c}%
\end{table}%

\noindent\textbf{Effect of the Weight $\lambda_c$.} Furthermore, we investigate the influence of varying weights ($\lambda_c$) for the confidence regularization loss on model performance. Table~\ref{tab:c} shows the pixel loss and perceptual loss of the model on the test set under different weights. We observe that when $\lambda_c >1$, both the pixel loss and perceptual loss of the model are relatively high, indicating poor image quality. When $\lambda_c <0.5$, the model exhibits relatively lower pixel loss, and when $\lambda_c =0.1$, the model achieves the minimum perceptual loss. Consequently, based on the experimental results, we set $\lambda_c =0.1$ as the default value.

\subsection{Further Discussion}

\begin{figure}[t]
\centering
    \includegraphics[width=0.48\textwidth]{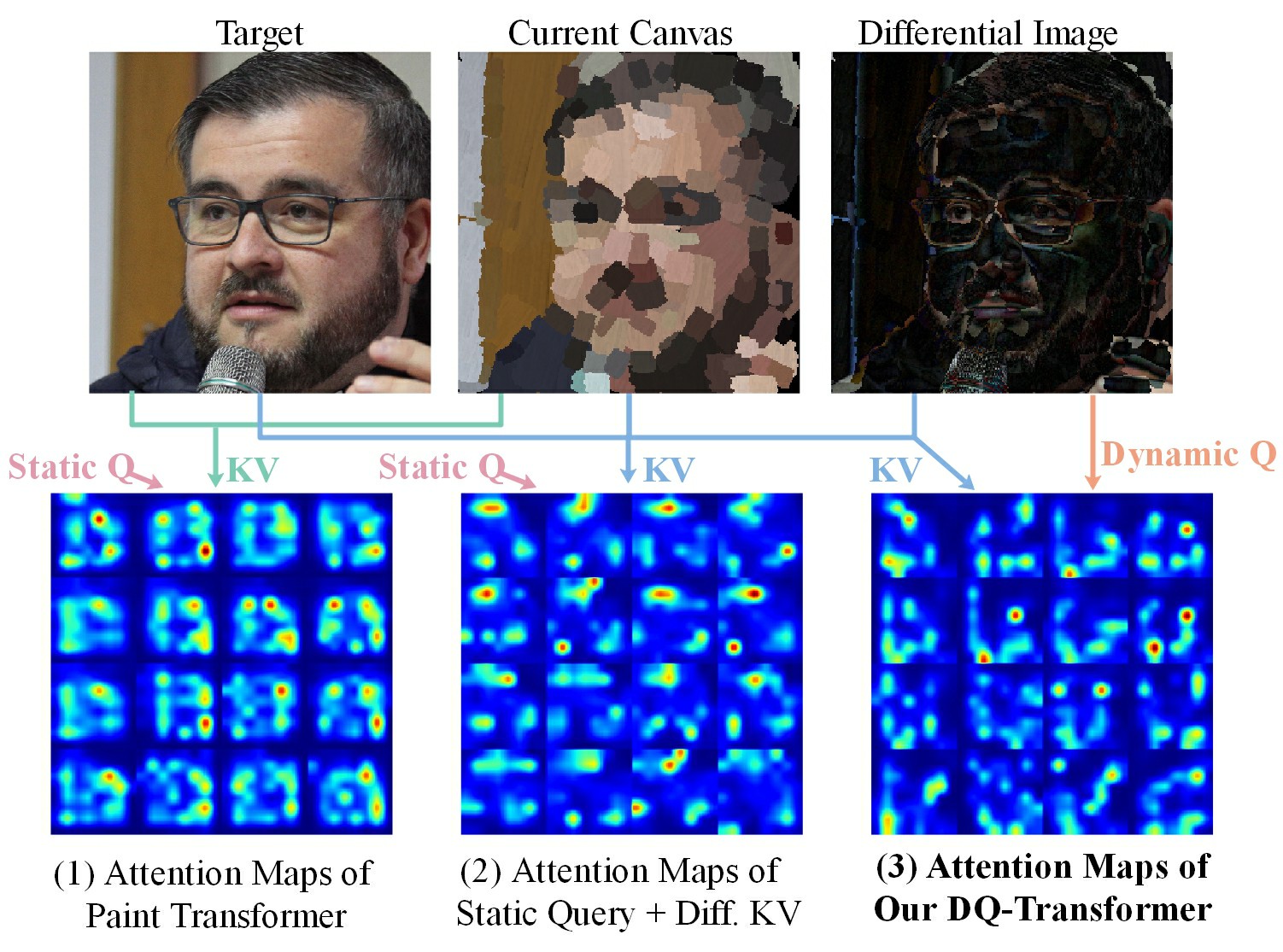}
    \caption{Attention visualization comparing three configurations. (1) Paint Transformer uses fixed learnable queries and key-value from the target and current canvas. (2) A variant with static queries and key-value from target, canvas, and differential image. (3) Our DQ-Transformer uses the differential image as dynamic query and combines all three inputs for key-value. Our model generates attention maps that sharply focus on regions with significant reconstruction errors.}
    \label{fig:att}
\end{figure}

\begin{figure}[t]
\centering
    \includegraphics[width=0.48\textwidth]{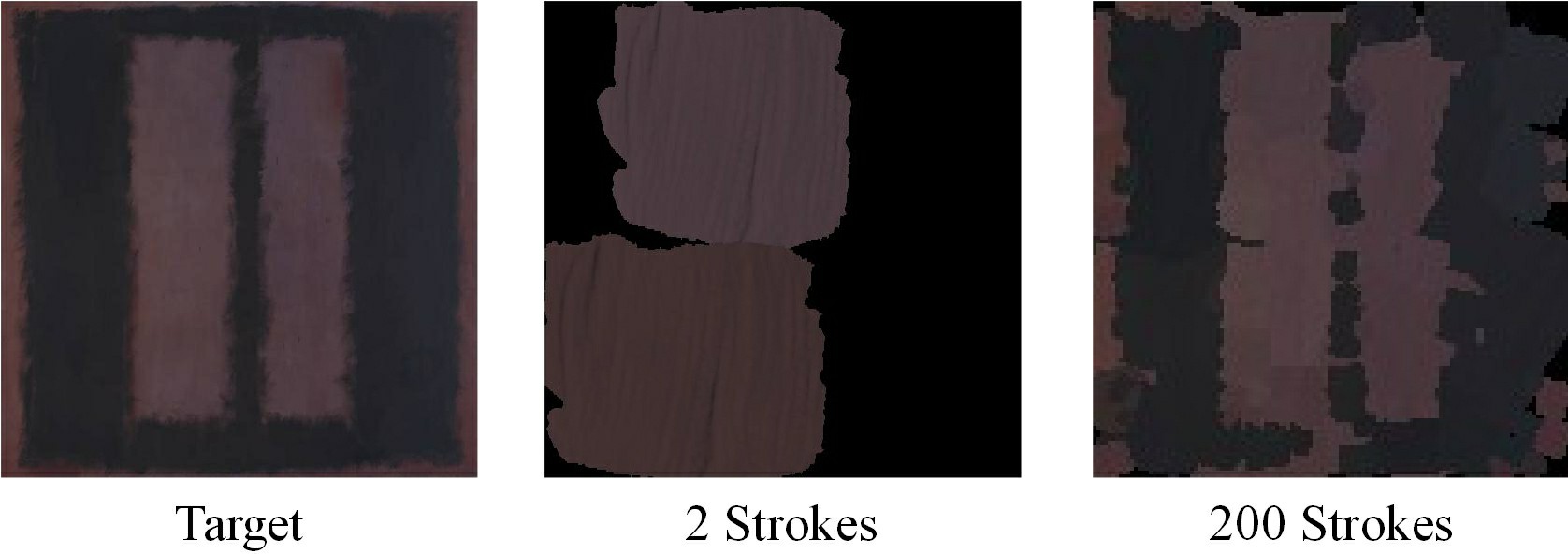}
    \caption{Failure case under an extremely limited stroke budget. With only two strokes, the model produces a highly abstract output that cannot capture the structure of the target image, illustrating a fundamental limitation of stroke-based neural painting methods.}
    \label{fig:fc}
\end{figure}

\noindent\textbf{Attention Analysis.} To better understand how our differential query mechanism influences stroke placement, we visualize cross-attention maps from three configurations: (1) the original Paint Transformer with fixed learnable queries; (2) a variant where the differential image is used as key-value but queries remain static; (3) our DQ-Transformer, where the differential image serves as dynamic queries. As shown in Figure~\ref{fig:att}, we visualize the cross-attention maps from the first decoder layer of transformer. It's important to note that the full-image attention map is stitched from 16 local maps, as the model processes the image in a 4×4 patch grid during inference. The Paint Transformer exhibits scattered attention patterns with no clear spatial correlation to reconstruction errors. When the differential image is used only as key-value, attention becomes slightly more focused but still fails to consistently highlight under-reconstructed regions. In contrast, our method produces sharp, localized attention peaks that align closely with high-error areas in the differential image. By formulating the differential image as a dynamic query, our model embodies the “look, compare, and draw” painting paradigm: it first looks at the current canvas, compares it with the target to compute residual errors, and then draws strokes guided by those discrepancies. This feedback-driven loop enables the model to allocate brushstrokes adaptively, focusing on regions that need refinement rather than applying uniform coverage.

\noindent\textbf{Comparison with General Image Stylization Methods.} Recent advances in diffusion models and large vision-language systems, such as StyleAligned~\cite{stylealigned} and B-LoRA~\cite{blora}, have achieved impressive results in global style transfer and semantic image manipulation. However, these methods operate in pixel or latent space and generate images holistically, without explicitly modeling the painting process. In contrast, our work falls within the neural painting paradigm, where the core objective is to learn stroke-level prediction through a coarse-to-fine autoregressive sequence. This formulation naturally yields a complete painting trajectory that can be rendered as a temporally coherent animation. Moreover, the generated stroke sequences and their intermediate renderings constitute high-quality, scalable training data for models that aim to learn from procedural creation dynamics. For instance, ProcessPainter~\cite{processpainter} leverages neural painting pipelines to synthesize video datasets capturing stroke-by-stroke artistic generation. Crucially, our method requires no real-world oil-painting images for training, relying solely on synthetic strokes, and thus avoids dependence on scarce artistic datasets.
Beyond data synthesis, our explicit stroke programs are directly executable by robotic painting systems. Each predicted stroke includes geometric and appearance parameters that can be translated into motor commands for physical brushes. While producing a finished artwork via printing or direct pixel rendering is technically straightforward, the ability to generate a painting step by step in real time, adapting brushstrokes based on ongoing canvas feedback, mirrors how humans create art and enables truly interactive and observable artistic behavior. This makes our approach particularly attractive for applications such as robot art education and human-robot co-creation.
Our method is not intended to replace general-purpose stylization tools, but rather to complement them by offering a process-driven approach to art generation.

\noindent\textbf{Limitations.} We acknowledge that our model shares a common limitation with other neural painting approaches. When restricted to an extremely small number of strokes, such as two, it cannot faithfully reconstruct the input image. As shown in Figure~\ref{fig:fc}, the output under this setting is highly abstract and lacks structural fidelity. This behavior stems from the nature of stroke based generation. Each brushstroke is a local and spatially constrained operation. With only a few strokes available, the model has insufficient capacity to represent complex shapes or fine details. It reflects a fundamental constraint of the neural painting paradigm, which relies on iterative refinement over many steps. As the stroke budget increases, for example to 200 strokes, the reconstruction quality improves significantly. Therefore, very low stroke counts should be understood as early sketching stages rather than final outputs. Future work will explore hybrid strategies that combine semantic priors, \eg, keypoints~\cite{lin2022joint}, with stroke based rendering to enhance early stage expressiveness.

\section{Conclusion}
In this work, we introduce a new automatic oil painting method guided by differential images, which generates brushstrokes akin to those created by human artists. We design a Differential Query Transformer and incorporate the differential image features as queries for decoding the brushstrokes. This ``Look, Compare and Draw'' approach enables the model to precisely focus on the visual effects produced by the incremental addition of strokes. Coupled with adversarial training, this mechanism significantly improves stroke prediction accuracy and, subsequently, enhances the fidelity of the output images.
We have conducted comparisons against state-of-the-art stroke-based painting methods on unseen real-world datasets and validated the superiority of our method through a combination of qualitative and quantitative evaluations, as well as a user study, assessing both pixel-level and perception-level reconstruction accuracy.

\bibliographystyle{IEEEtran}
\bibliography{main}

\end{document}